\documentclass[letterpaper]{article} 

\usepackage{aaai2027}
\usepackage{times}        
\usepackage{helvet}       
\usepackage{courier}      
\usepackage[hyphens]{url} 
\usepackage{graphicx}     
\usepackage{natbib}       
\usepackage{caption}      
\usepackage{amsmath,amssymb,mathtools}

\usepackage{booktabs}
\usepackage{multirow}
\usepackage{makecell}
\usepackage{tabularx}
\usepackage{array}
\usepackage{ragged2e}
\usepackage[table]{xcolor}

\usepackage{tikz}
\usetikzlibrary{arrows.meta, positioning, fit}

\usepackage{subcaption}

\usepackage{enumitem}
\usepackage{soul}
\usepackage{microtype}

\usepackage{algorithm}
\usepackage{algpseudocode}

\usepackage{float}
\usepackage{placeins}
\usepackage{newfloat}

\usepackage{bbding}
\usepackage{pifont}

\newcommand{\cmark}{\ding{51}}
\newcommand{\xmark}{\ding{55}}

\usepackage{xspace}
\usepackage{xstring}

\usepackage{listings}

\DeclareCaptionStyle{ruled}{labelfont=normalfont,labelsep=colon,strut=off} 

\floatstyle{ruled}
\newfloat{listing}{tb}{lst}{}
\floatname{listing}{Listing}

\setlist{topsep=2pt,itemsep=1pt,parsep=0pt,partopsep=0pt}

\definecolor{accentorange}{RGB}{230,126,34}
\definecolor{accentblue}{RGB}{31,119,180}

\newcommand{\benchmark}{\texttt{FAME}\xspace}

\definecolor{famspec}{RGB}{78,121,167}   
\definecolor{famsam}{RGB}{242,142,43}    
\definecolor{famclip}{RGB}{89,161,79}    
\definecolor{fammllm}{RGB}{176,122,161}  
\definecolor{famsup}{RGB}{225,87,89}     
\newcommand{\fs}{\textsuperscript{\textcolor{famspec}{$\blacklozenge$}}}     
\newcommand{\fsam}{\textsuperscript{\textcolor{famsam}{$\bullet$}}}          
\newcommand{\fclip}{\textsuperscript{\textcolor{famclip}{$\blacktriangle$}}} 
\newcommand{\fmllm}{\textsuperscript{\textcolor{fammllm}{$\blacksquare$}}}   
\newcommand{\teststat}[3]{%
  \makebox[2.8em][r]{#1}%
  \hspace{0.4em}%
  \makebox[5.5em][l]{(#2, #3)}%
}

\newcommand{\papertitle}{
Benchmarking Foundation and Large Language Models for Few-Shot Medical Image Segmentation
}

\title{\papertitle}

\author{
    Jinghong Liu\equalcontrib,
    Yuchuan Deng\equalcontrib,
    Fanping Liu,
    Meng Huang,
    Xirong Li
    \thanks{
        Corresponding author: xirong@ruc.edu.cn
    }
}

\affiliations{
    Renmin University of China
}

\nocopyright
\begin{document}

\maketitle

\begin{abstract}
Few-shot medical image segmentation (FS-MIS) aims to segment novel regions of interest (ROIs) from a few annotated support examples. 
Despite rapid progress, existing FS-MIS solutions span diverse paradigms but are evaluated under inconsistent settings, leaving their relative effectiveness unclear.
We introduce \benchmark, a unified benchmark for evaluating FS-MIS solutions, covering specialists, SAM-based methods, CLIP-based methods, and MLLM-based methods. 
\benchmark contains 14,958 test samples across 7 anatomical sites, 9 imaging modalities, and 14 ROI categories, and evaluates models under zero-shot and ten-shot settings with additional assessment of target-absence recognition and generalization under covariate and semantic shifts.
Our evaluation reveals several findings. 
First, effective few-shot segmentation depends on how models exploit support examples: direct visual adaptation generally outperforms prompt-based strategies. 
Second, increasing support examples improves performance only when models can effectively utilize them. 
Third, semantic transfer remains substantially more challenging than imaging-domain adaptation, and strong localization ability does not necessarily imply reliable target-absence recognition.
We hope \benchmark provides a comprehensive understanding of current FS-MIS solutions and facilitates the development of more effective and reliable few-shot medical segmentation methods.
\end{abstract}

\begin{links}
    \link{Code}{https://github.com/Kimokcheon/FAME}
    \link{Dataset}{https://huggingface.co/datasets/Kimokcheon/FAME_benchmark}
    \link{Project Page}{https://rucerchui.github.io/FAME}
\end{links}

\section{Introduction}

\begin{figure}[!t]
\centering

\begin{subfigure}[t]{\linewidth}
\centering
\includegraphics[width=\linewidth]{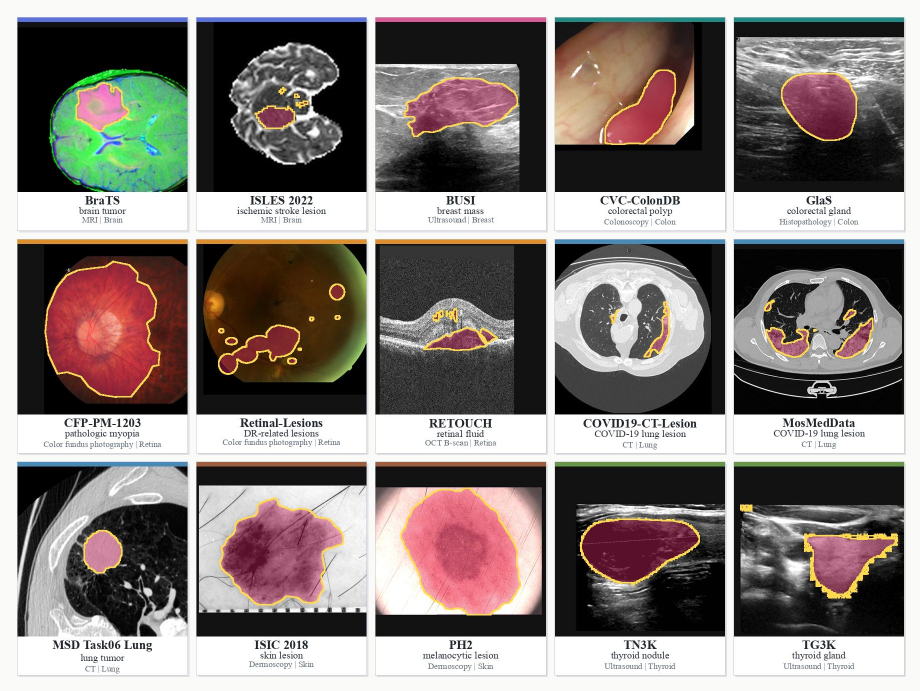}
\caption{
Region-of-interest (ROI) covered by \benchmark.
}
\label{fig:fame_targets}
\end{subfigure}

\begin{subfigure}[t]{\linewidth}
\centering
\includegraphics[width=\linewidth]{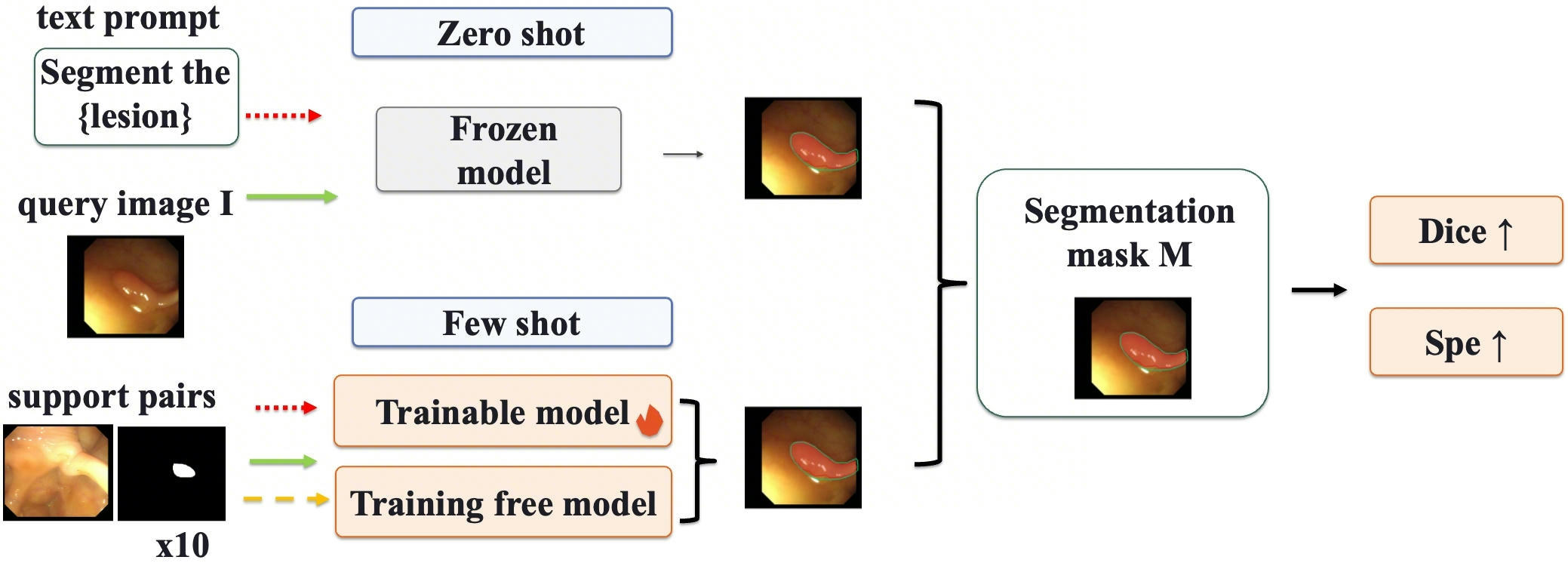}
\caption{Zero-shot and few-shot evaluation modes.}
\label{fig:fame_modes}
\end{subfigure}

\begin{subfigure}[t]{\linewidth}
\centering
\includegraphics[width=\linewidth]{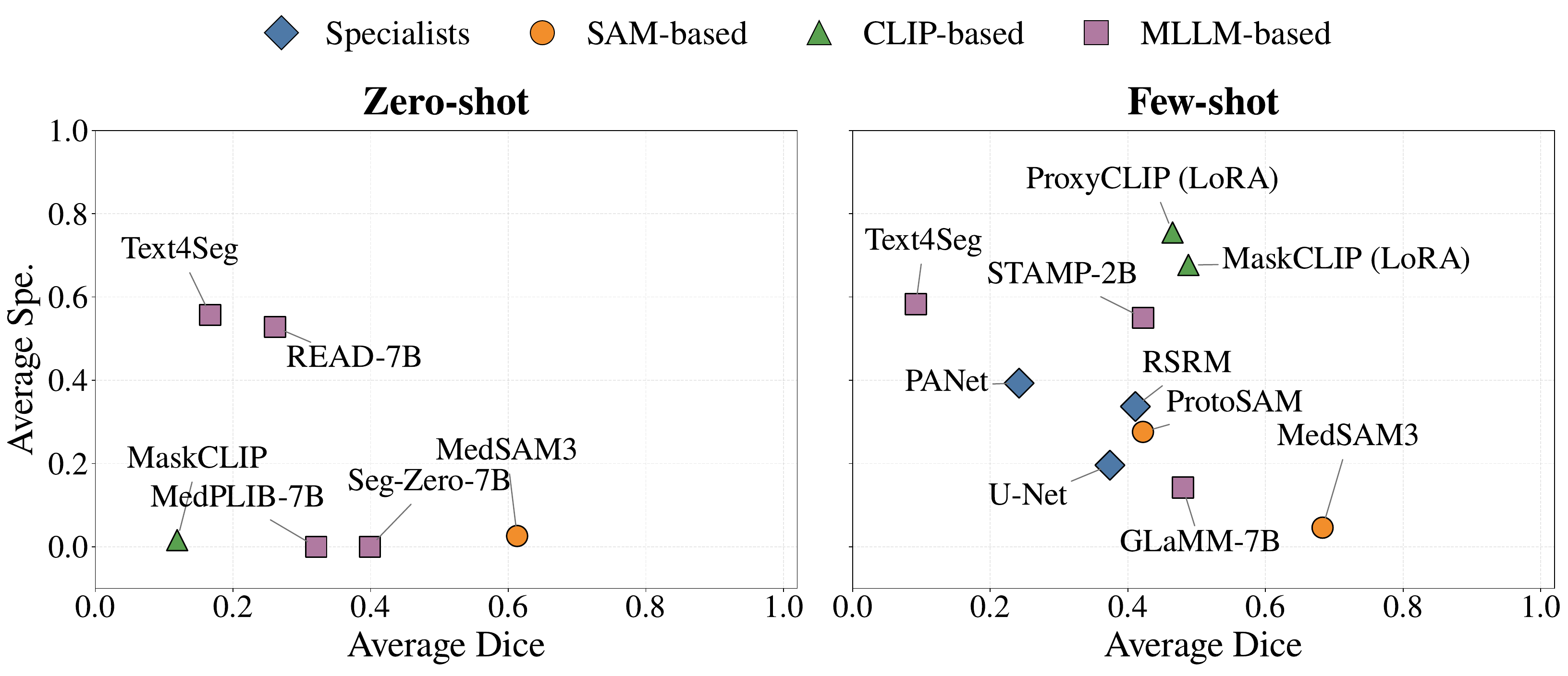}
\caption{Dice--specificity trade-off.}
\label{fig:fame_performance}
\end{subfigure}

\caption{\textbf{Overview of the proposed \benchmark benchmark for few-shot medical image segmentation (FS-MIS)}. 
}
\label{fig:fame_overview}
\end{figure}


Few-shot medical image segmentation (FS-MIS) aims to segment a region of interest (ROI), such as an anatomical structure or lesion, in a \emph{query image} using only a few \emph{support images} with pixel-level masks. 
Support examples provide limited visual and spatial cues, which the model must exploit to segment the ROI in query images~\cite{dissanayake2025fewshot}.

\begin{table*}[t]
\centering
\setlength{\tabcolsep}{4pt}
\renewcommand{\arraystretch}{1.5}
\resizebox{\textwidth}{!}{
\begin{tabular}{@{}lcccccccc@{}}
\toprule
\textbf{MIS Task} 
& \makecell{\textbf{ChatGPT}\\[-1mm]\cite{openai2022chatgpt}}
& \makecell{\textbf{Claude}\\[-1mm]\cite{anthropic2023claude}}
& \makecell{\textbf{DeepSeek}\\[-1mm]\cite{deepseek2024chat}}
& \makecell{\textbf{Doubao}\\[-1mm]\cite{bytedance2023doubao}}
& \makecell{\textbf{Gemini}\\[-1mm]\cite{google2024gemini}}
& \makecell{\textbf{GLM}\\[-1mm]\cite{zhipu2023chatglm}}
& \makecell{\textbf{Kimi}\\[-1mm]\cite{moonshot2023kimi}}
& \makecell{\textbf{Qwen}\\[-1mm]\cite{qwen2023assistant}} \\
\midrule
\makecell{Brain tumor segmentation\\ in MRI images}  & \makecell{MedSAM\fsam\\\scriptsize LoRA} & \makecell{U-Net\fs\\\scriptsize FPT} & \makecell{U-Net\fs\\\scriptsize FPT} & \makecell{MedSAM3\fsam\\\scriptsize LoRA} & \makecell{MedSAM\fsam\\\scriptsize Interface} & \makecell{MedSAM3\fsam\\\scriptsize LoRA} & \makecell{MedSAM3\fsam\\\scriptsize Interface} & \makecell{U-Net\fs\\\scriptsize FPT} \\

\midrule

\makecell{Colon polyp segmentation\\ in endoscopy images}     & \makecell{SAM\fsam\\\scriptsize Interface} & \makecell{MedSAM\fsam\\\scriptsize LoRA} & \makecell{U-Net\fs\\\scriptsize FPT} & \makecell{MedSAM3\fsam\\\scriptsize LoRA} & \makecell{LISA\fmllm\\\scriptsize LoRA} & \makecell{MedPLIB\fmllm\\\scriptsize LoRA} & \makecell{MedSAM\fsam\\\scriptsize Interface} & \makecell{SAM\fsam\\\scriptsize LoRA} \\

\midrule

\makecell{Pathologic myopia segmentation\\ in color fundus photos}   & \makecell{MedSAM\fsam\\\scriptsize LoRA} & \makecell{U-Net\fs\\\scriptsize FPT} & \makecell{MedSAM\fsam\\\scriptsize LoRA} & \makecell{SCLIP\fclip\\\scriptsize Interface} & \makecell{LISA\fmllm\\\scriptsize LoRA} & \makecell{MedPLIB\fmllm\\\scriptsize LoRA} & \makecell{ProxyCLIP\fclip\\\scriptsize Interface} & \makecell{SAM\fsam\\\scriptsize LoRA} \\

\midrule

\makecell{Skin lesion segmentation\\ in dermoscopy images}     & \makecell{SAM\fsam\\\scriptsize Interface} & \makecell{MedSAM\fsam\\\scriptsize LoRA} & \makecell{U-Net\fs\\\scriptsize FPT} & \makecell{MaskCLIP\fclip\\\scriptsize Interface} & \makecell{MedSAM\fsam\\\scriptsize Interface} & \makecell{MedSAM3\fsam\\\scriptsize LoRA} & \makecell{SAM\fsam\\\scriptsize LoRA} & \makecell{SAM\fsam\\\scriptsize LoRA} \\
\bottomrule
\end{tabular}}
\caption{ \textbf{AI-recommended solutions for FS-MIS}. 
Given a novel MIS task defined by a handful of support images, the text below each method describes how to adapt the method: \emph{LoRA} (low-rank adaptation), \emph{FPT} (full-parameter training from scratch), and \emph{Interface} (training only a segmentation interface with the backbone frozen). We categorize the AI-recommended methods into four categories: \emph{SAM}-based\fsam (SAM, MedSAM, MedSAM3), \emph{CLIP}-based\fclip (SCLIP, MaskCLIP, ProxyCLIP), \emph{MLLM}-based\fmllm (LISA, MedPLIB) and \emph{Specialists}\fs (U-Net). By evaluating 23 methods on 14 FS-MIS tasks with a unified evaluation protocol, the proposed \benchmark benchmark aims to answer which solution is the choice.  
}
\label{tab:llm_survey}
\end{table*}

Recent FS-MIS solutions span several methodological paradigms. 
One line of work develops specialist segmentation models that directly leverage support examples for query prediction, either by matching query features with support-derived prototypes~\cite{wang2019panet,sun2026training} or by learning richer support--query interactions across segmentation tasks~\cite{butoi2023universeg,rakic2024tyche,bo2025famnet}. 
Another line adapts foundation models to FS-MIS. Promptable segmentation models convert annotated support images into spatial or feature-level prompts, thereby reusing pretrained mask-generation priors for unseen medical ROIs~\cite{xu2024sammpa,ayzenberg2025protosam,bo2026focus}. 
Recent studies have further explored leveraging the broad pretrained knowledge of MLLMs for FS-MIS~\cite{medplib}.

However, these solutions are typically evaluated under isolated few-shot settings, including different numbers of support pairs, imaging modalities, and ROIs, making it impossible to determine which paradigm is preferable under a common FS-MIS setting.
Consequently, it remains unclear which solution should be selected for a practical FS-MIS scenario.
To illustrate this challenge, consider a practitioner who has a few medical images with masks for a target ROI and seeks a model that can segment the same ROI in query images.
A natural first step is to consult an AI assistant for a suitable technical solution.
As an informal probe, we prompted eight publicly accessible AI assistants.
Their recommendations varied substantially across assistants and tasks (see Tab.~\ref{tab:llm_survey}), highlighting a key question: \emph{which solution is most effective for few-shot medical image segmentation?}

Existing benchmarks do not comprehensively compare these solutions under a unified few-shot setting.
A recent survey~\cite{dissanayake2025fewshot} and UniMedDB~\cite{zhao2026segmic} mainly evaluate prototype-matching-based training-free specialist methods, while MeCoVQA-S~\cite{shen2026image} focuses on MLLM-based segmentation models. 
These evaluations are limited to individual solution categories and primarily assess positive-case segmentation accuracy, overlooking the ability to correctly reject absent ROIs.

To address this gap, we present \benchmark, a unified benchmark for \textbf{F}ound\textbf{A}tion and large language models for few-shot \textbf{M}edical image s\textbf{E}gmentation, including specialists, SAM-based methods, CLIP-based methods, and MLLM-based methods. 
\benchmark comprises 14,958 test samples spanning 7 anatomical sites, 9 imaging modalities, and 14 classes of ROIs (see Fig.~\ref{fig:fame_targets}).
Per task we also include negative examples when available in the test set, to assess to what extent a model makes false alarms. 
To investigate how different architectures and adaptation strategies affect FS-MIS performance, we evaluate these solutions under two complementary settings. The \emph{Zero-shot} setting measures their segmentation ability without task-specific examples, while the \emph{Few-shot} setting evaluates their ability to leverage limited support examples for target-specific segmentation (see Fig.~\ref{fig:fame_modes}).Beyond these two settings, we further test how well each solution generalizes when the test images differ from the support images, separating covariate shift (the same ROI under new imaging conditions) from semantic shift (a new, previously unseen ROI).

Our evaluation reveals several challenges in existing FS-MIS solutions. 
First, effective adaptation depends on how models exploit support masks: incorporating them directly into visual representations or mask generation outperforms indirect prompt or semantic adaptation (Fig.~\ref{fig:fame_performance}). 
Second, adding support examples helps only when models can utilize them. 
Finally, few-shot segmentation shows a clear gap between localization and target-absence recognition, and transferring to unseen ROIs is far harder than handling imaging-domain shifts.

The main contributions are summarized as follows:
\begin{itemize}
\item \textbf{Benchmark.} We introduce \benchmark, a comprehensive benchmark for FS-MIS, covering 14,958 test samples spanning 7 anatomical sites, 9 imaging modalities, and 14 classes of ROIs.
\benchmark provides a unified evaluation protocol that evaluates pretrained models in both zero-shot settings and after ten-shot adaptation, with additional assessment of generalization under covariate and semantic shifts.

\item \textbf{Evaluation.} We evaluate four solution categories along with their corresponding adaptation strategies.

\item \textbf{Findings.} We reveal that effective few-shot medical segmentation depends on how models exploit support examples, with direct visual adaptation generally outperforming prompt-based strategies. 
We further show that support scaling is effective only when models can utilize additional examples, while semantic transfer and reliable target-absence recognition remain major challenges.

\end{itemize}

\section{Related Work}


\begin{table*}[t]
\centering
\small
\setlength{\tabcolsep}{5pt}
\renewcommand{\arraystretch}{0.8}

\begin{tabularx}{\linewidth}{
    @{}
    l
    l
    l
    l
    l
    @{}
}
\toprule
\textbf{Source Dataset}
& \textbf{Anatomy}
& \textbf{Image modality}
& \textbf{Region-of-Interest}
& \arraybackslash\textbf{Test Samples} \\
\midrule

BraTS~\cite{menze2015multimodal}
& Brain
& MRI
& Brain tumor
& \teststat{299}{237}{62} \\

ISLES 2022~\cite{hernandez2022isles}
& Brain
& MRI
& Ischemic stroke lesion
& \teststat{1,000}{700}{300} \\

BUID~\cite{abbasian2023open}
& Breast
& Ultrasound
& Breast mass
& \teststat{222}{222}{0} \\

BUSI~\cite{aldhabyani2020dataset}
& Breast
& Ultrasound
& Breast mass
& \teststat{770}{637}{133} \\

CVC-ColonDB~\cite{bernal2012towards}
& Colon
& Colonoscopy
& Colorectal polyp
& \teststat{370}{370}{0} \\

CVC-ClinicDB~\cite{bernal2015wm}
& Colon
& Colonoscopy
& Colorectal polyp
& \teststat{602}{602}{0} \\

GlaS~\cite{SIRINUKUNWATTANA2017489}
& Colon
& Histopathology
& Colorectal gland
& \teststat{770}{637}{133} \\

Kvasir-SEG~\cite{jha2020kvasirseg}
& Colon
& Colonoscopy
& Colorectal polyp
& \teststat{990}{990}{0} \\

OIA-DDR~\cite{LI2019}
& Retina
& Color fundus photography
& DR-related lesions
& \teststat{747}{747}{0} \\

CFP-PM-1203~\cite{tang2022artificial}
& Retina
& Color fundus photography
& Pathologic myopia
& \teststat{1,000}{959}{41} \\

Retinal-Lesions~\cite{icpr2020-LesionNet}
& Retina
& Color fundus photography
& DR-related lesions
& \teststat{1,000}{700}{300} \\

RETOUCH~\cite{bogunovic2019retouch}
& Retina
& OCT B-scan
& Retinal fluid
& \teststat{1,000}{700}{300} \\

COVID19-CT-Lesion~\cite{maftouni2021robust}
& Lung
& CT
& COVID-19 lung lesion
& \teststat{999}{999}{0} \\

MosMedData~\cite{morozov2020mosmeddata}
& Lung
& CT
& COVID-19 lung lesion
& \teststat{999}{999}{0} \\

MSD Task06 Lung~\cite{antonelli2022medicaldecathlon}
& Lung
& CT
& Lung tumor
& \teststat{1,000}{1,000}{0} \\

PH2~\cite{mendonca2013ph2}
& Skin
& Dermoscopy
& Melanocytic lesion
& \teststat{190}{190}{0} \\

ISIC 2018~\cite{codella2019skin}
& Skin
& Dermoscopy
& Skin lesion
& \teststat{1,000}{1,000}{0} \\

TG3K~\cite{gong2023thyroid}
& Thyroid
& Ultrasound
& Thyroid gland
& \teststat{1,000}{1,000}{0} \\

TN3K~\cite{gong2021multi-task}
& Thyroid
& Ultrasound
& Thyroid nodule
& \teststat{1,000}{1,000}{0} \\

\bottomrule
\end{tabularx}

\caption{
\textbf{Data sources of \benchmark}.
Per test set, numbers in parentheses are the amount of positive / negative samples.
}
\label{tab:source_datasets}
\end{table*}

Few-shot medical image segmentation has witnessed rapid progress, with diverse solution paradigms emerging; however, existing evaluations remain fragmented across approaches and experimental settings.

Existing evaluations related to FS-MIS mainly investigate individual solution categories. 
UniMedDB~\cite{zhao2026segmic} evaluates universal in-context segmentation models across medical domains, while MeCoVQA-S~\cite{shen2026image} focuses on the segmentation capability of medical MLLMs. 
The survey~\cite{dissanayake2025fewshot} further summarizes recent FS-MIS advances and compares representative prototype-based and meta-learning approaches. 
Although these studies provide valuable insights into specific settings, they differ in support configurations, datasets, target ROIs, and evaluation protocols, making direct comparison across different solution paradigms difficult. 
Moreover, existing evaluations mainly emphasize positive-case segmentation accuracy, leaving the ability to recognize absent ROIs insufficiently studied.

Beyond evaluation protocols, existing FS-MIS methods have also evolved along multiple directions according to how support examples are utilized. 
Early specialists~\cite{wang2019panet,butoi2023universeg,rakic2024tyche,bo2025famnet,sun2026training} are specifically designed for support-conditioned segmentation, learning target-specific representations or support--query interactions from few-shot examples. 
With the emergence of foundation models, recent approaches adapt pretrained models to FS-MIS. 
SAM-based methods~\cite{xu2024sammpa,ayzenberg2025protosam,bo2026focus} leverage promptable segmentation priors, CLIP-based methods exploit vision-language representations for semantic guidance, and MLLM-based approaches~\cite{medplib} extend multimodal large language models with segmentation capabilities.

Despite these advances, existing evaluations do not provide a comprehensive comparison among specialists, foundation-model-based, and MLLM-based methods under a common FS-MIS protocol. 
\benchmark addresses these limitations by evaluating diverse solution paradigms across heterogeneous medical targets, incorporates both positive and negative query cases under both zero-shot and few-shot settings.

\section{Our Roadmap to \benchmark}
\label{sec:proposed_benchmark}
This section presents \benchmark from two aspects.
Sec.~\ref{sec:benchmark_construction} describes the construction of \benchmark.
Sec.~\ref{sec:evaluation_protocol} introduces the evaluation protocol.

\subsection{Dataset Construction}
\label{sec:benchmark_construction}

\benchmark organizes diverse medical segmentation datasets into task-specific region-of-interest (ROI) segmentation tasks. 
Each task corresponds to a clinically meaningful ROI concept and is designed to evaluate whether models can identify and segment the corresponding ROI from medical images under different supervision settings.

We collect 19 publicly available medical segmentation datasets, covering diverse organs, imaging modalities, and pathological conditions, as summarized in Tab.~\ref{tab:source_datasets}. 
To unify heterogeneous datasets, we represent each sample as an image--ROI-mask triplet $(I,R,M)$, where $I$ denotes the medical image, $R$ specifies the ROI concept, and $M$ denotes the corresponding binary segmentation mask.

For datasets containing multiple annotations that jointly characterize the same clinical condition, we merge these annotations into a unified ROI mask through pixel-wise union. 
This design follows the objective of \benchmark, which focuses on ROI-level segmentation capability rather than fine-grained pathological subtype classification. 
For example, diabetic-retinopathy-related findings, including microaneurysms, hard exudates, and neovascularization, represent different manifestations of the same pathological condition and are therefore combined into a single ROI mask. 
A pixel is labeled as foreground if it belongs to any associated annotation and background otherwise.

For each ROI task, a sample is considered \emph{positive} if its ROI mask contains at least one foreground pixel, indicating the presence of the ROI. 
Otherwise, it is considered \emph{negative}. 
Negative samples are retained whenever available to evaluate target-absence recognition.

We retain only diagnostically usable images and apply standardized quality control to all image--mask pairs. 
Samples with invalid annotations, mismatched image--mask pairs, malformed masks, or severe quality issues are removed.

After quality control, we perform content-based deduplication to remove repeated or near-identical images. 
For each ROI task, we randomly select 50 unique positive samples to construct the support pool. 
The support pool is further divided into five disjoint 10-shot support sets for evaluation. 
All remaining samples are reserved as test samples. 
The support and test sets are strictly disjoint, with no duplicated images, and test annotations are never accessed during model adaptation.
Ultimately, \benchmark contains 950 support samples and 14,958 test samples organized into task-specific ROI segmentation tasks.

\subsection{Evaluation Protocol}
\label{sec:evaluation_protocol}

\subsubsection{Evaluation Modes}
\label{sec:evaluation_settings}

\benchmark defines two complementary evaluation modes: \emph{zero-shot} mode and \emph{few-shot} mode, see Fig.~\ref{fig:fame_modes}.

In the \textbf{\textit{zero-shot mode}}, models receive only an image $I$ and a textual ROI concept $R$. 
The model directly predicts the binary ROI mask $\hat{M}$, evaluating the ROI recognition and segmentation capability acquired during pretraining. 
Each model is provided with the ROI concept using a unified prompt template, or its official prompt format when required.

In the \textbf{\textit{few-shot mode}}, each ROI task is evaluated through an independent adaptation episode. 
Each task provides five disjoint 10-shot support sets (Sec.~\ref{sec:benchmark_construction}). 
For each support split, the model is adapted using only the corresponding annotated image--mask pairs, and then evaluated on the held-out test samples. 
The final performance is the average across the five support splits, reducing the variance caused by support-set selection. Crucially, \benchmark fixes only the annotation budget (ten labeled pairs per task) and lets each method use it in its own way: some update parameters while others adapt without any parameter update. We treat this difference as an intrinsic property of each method rather than something to be equalized, matching the real choice a practitioner faces under a fixed labeling effort.

Beyond the standard in-distribution (IID) evaluation, where support and test samples share the same ROI task distribution, we evaluate out-of-distribution (OOD) generalization, where a model is adapted on a source task and tested on a different one~\cite{sun2026training}.
Following~\cite{yang2024generalized}, we consider two types of distribution shifts.
\emph{Covariate shift} preserves the ROI concept but changes the source dataset, evaluating robustness to variations in acquisition protocols and image appearance.
\emph{Semantic shift} changes the ROI concept between adaptation and test while keeping the imaging modality consistent whenever possible: a model adapted on a source concept is tested on a related target concept, which is given by the prompt but without any target support (Tab.~\ref{tab:oodse}), measuring whether the knowledge learned on the source concept transfers to the new target concept.
We therefore report OOD only for methods that are adapted on the source by updating parameters. Methods that instead condition on the support set at test time without updating any parameters have nothing to transfer and are excluded.

\begin{table}[t]
\centering
\renewcommand{\arraystretch}{0.8}
\small
\begin{tabular}{ll}
\toprule
\textbf{Source domain} & \textbf{Target domain} \\
\midrule
\multicolumn{2}{l}{\textit{Covariate shift}} \\
BUSI & BUID \\
Kvasir-SEG & CVC-ClinicDB \\
Kvasir-SEG & CVC-ColonDB \\
ISIC 2018 & PH2 \\
Retinal-Lesions & DDR \\
COVID19-CT-Lesion & MosMedData \\
\midrule
\multicolumn{2}{l}{\textit{Semantic shift}} \\
BraTS & ISLES 2022 \\
CFP-PM-1203 & Retinal-Lesions \\
CFP-PM-1203 & DDR \\
COVID19-CT-Lesion & MSD Task06 Lung \\
BUSI & TN3K \\
TN3K & TG3K \\
\bottomrule
\end{tabular}
\caption{\textbf{Two setups of out-of-distribution (OOD) evaluation}: covariate shift and semantic shift. 
}
\label{tab:oodse}
\end{table}

\subsubsection{Performance Metrics}
\label{sec:evaluation_metrics}
We use the same metrics for zero-shot and few-shot evaluation. 
For positive samples, we report Dice between predicted and ground-truth masks. 
For negative samples, we report image-level specificity (Spe.), defined as the proportion of ROI-absent images for which the model predicts an empty mask. 
Specificity is only evaluated on tasks containing negative samples, while the lung, skin, and thyroid tasks are evaluated only by Dice due to the absence of negative cases.

\subsubsection{Model Applicability}
\label{sec:model_applicability}

The few-shot mode applies to models that can be adapted using annotated image--mask pairs, whereas the zero-shot mode requires models to directly perform text-conditioned segmentation from an image and an ROI concept. 
Models supporting both text-conditioned segmentation and task-specific adaptation, such as LISA~\cite{lai2024lisa} and STAMP~\cite{liu2026stamp}, are evaluated in both modes. 
Models that only support adaptation from annotated image--mask pairs, such as F-LMM~\cite{wu2025flmmgroundingfrozenlarge} and LENS~\cite{liu2025segmentationplugandplaycapabilityfrozen}, are evaluated only in the few-shot mode.

\section{Methods Evaluated by \benchmark}
\label{sec:compared_methods}

\begin{table}[t]
\centering
\small
\setlength{\tabcolsep}{4pt}
\renewcommand{\arraystretch}{1}
\resizebox{\linewidth}{!}{%
\begin{tabular}{llcl}
\toprule
\textbf{Method}
& \textbf{Training components}
& \textbf{ZS}
& \textbf{Adaptation} \\

\midrule
\multicolumn{4}{l}{\textit{SAM-based segmentation models}}\\
SAM \textcolor{gray}{ICCV'23}
& --
& \xmark
& SAM-MPA \\

MedSAM \textcolor{gray}{Nat. Commun.'24}
& --
& \xmark
& SAM-MPA \\

MedSAM3 \textcolor{gray}{arXiv'25}
& SAM3 LoRA modules
& \cmark
& Official\\

ProtoSAM \textcolor{gray}{Sci. Rep.'25}
& --
& \xmark
& Official \\

FoB-SAM \textcolor{gray}{CVPR'26}
& --
& \xmark
& Official \\

\midrule
\multicolumn{4}{l}{\textit{CLIP-based dense localization models}}\\
SCLIP \textcolor{gray}{ECCV'24}
& Dense text--pixel similarity
& \cmark
& CoOp / LoRA \\

MaskCLIP \textcolor{gray}{ECCV'22}
& Dense text--pixel similarity
& \cmark
& CoOp / LoRA \\

ProxyCLIP \textcolor{gray}{ECCV'24}
& VFM-corrected dense similarity
& \cmark
& CoOp / LoRA \\

\midrule
\multicolumn{4}{l}{\textit{Burned-in MLLMs}}\\
LISA \textcolor{gray}{CVPR'24}
& MLLM projections; SEG MLP; decoder
& \cmark
& Official \\

GLaMM \textcolor{gray}{CVPR'24}
& Text MLP;  decoder
& \cmark
& LISA \\

READ\textcolor{gray}{CVPR'25}
& MLLM projections; SEG MLP; decoder
& \cmark
& Official \\

MedPLIB \textcolor{gray}{AAAI'25}
& Text MLP;  decoder
& \cmark
&  Official \\

Text4Seg \textcolor{gray}{ICLR'25}
& mm-projector ; MLLM language layers
& \cmark
& Official \\


STAMP \textcolor{gray}{CVPR'26}
& MLLM linear layers; mask-token classifier
& \cmark
& Official \\

\midrule
\multicolumn{4}{l}{\textit{Plug-in MLLMs}}\\
F-LMM \textcolor{gray}{CVPR'25}
& Attention-to-mask U-Net and SAM refiner
& \cmark
& Official
\\

LENS \textcolor{gray}{arXiv'25}
& Keypoint-prompt head and SAM decoder
& \cmark
& Official
\\

\midrule
\multicolumn{4}{l}{\textit{Specialists}}\\
U-Net \textcolor{gray}{MICCAI'15}
& Full Network
& \xmark
& FPT \\

PANet \textcolor{gray}{ICCV'19}
& --
& \xmark
& Training-free \\

UniverSeg \textcolor{gray}{ICCV'23}
& --
& \xmark
& Training-free \\

Tyche \textcolor{gray}{CVPR'24}
& --
& \xmark
& Training-free \\

FAMNet \textcolor{gray}{AAAI'25}
& --
& \xmark
& Training-free \\

RSRM \textcolor{gray}{ECCV'26}
& --
& \xmark
& Training-free \\

\bottomrule
\end{tabular}%
}
\caption{
Evaluated methods and their adaptation strategies in \benchmark. ZS denotes zero-shot segmentation using an image and a textual prompt.
}
\label{tab:compared_methods}
\end{table}

\begin{figure}[!t]
\centering
\begin{subfigure}[t]{\linewidth}
\centering
\includegraphics[width=\linewidth]{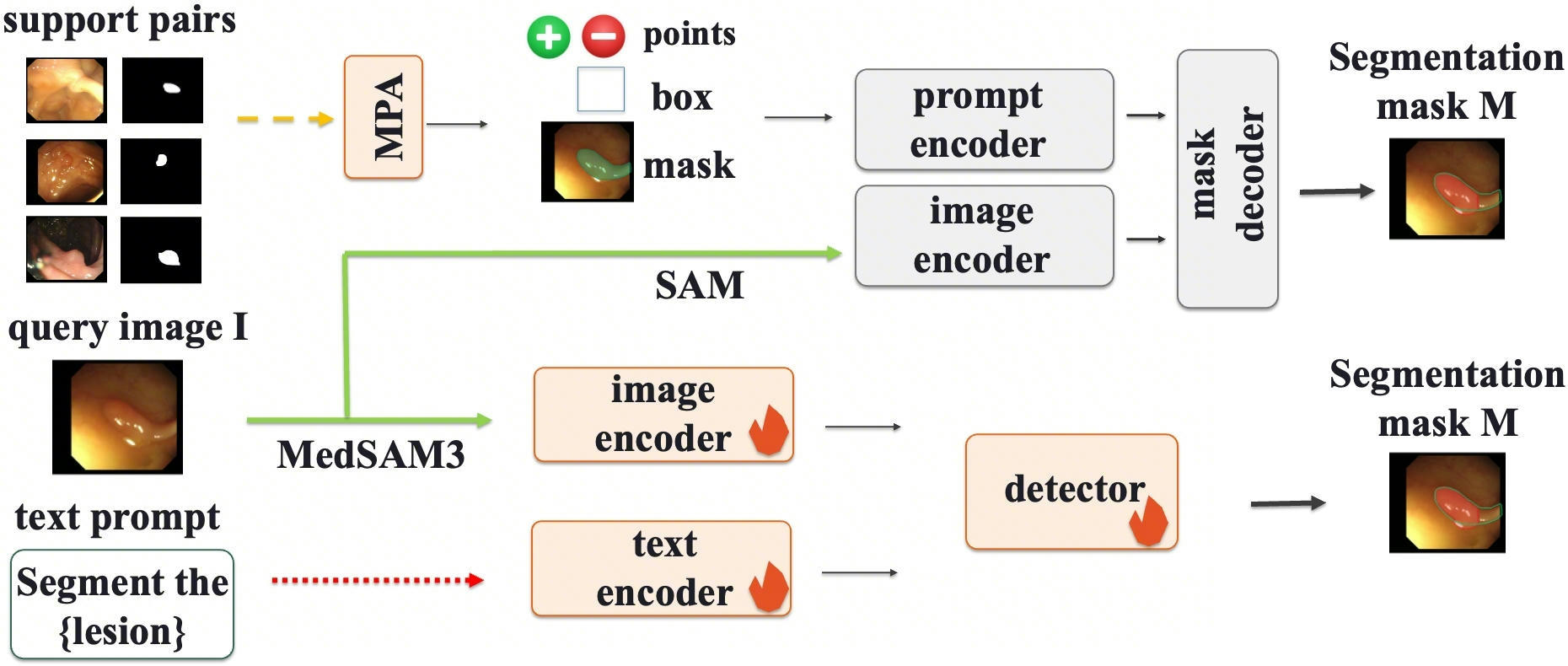}
\caption{SAM-based.}
\label{fig:gallery_sam}
\end{subfigure}

\vskip 0.5\baselineskip

\begin{subfigure}[t]{\linewidth}
\centering
\includegraphics[width=\linewidth]{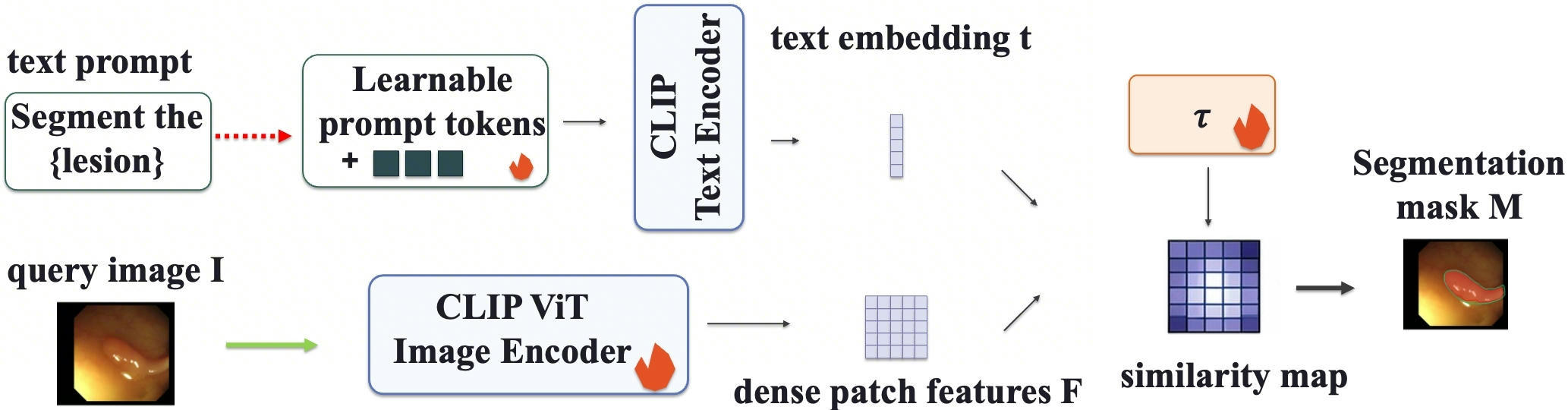}
\caption{CLIP-based.}
\label{fig:gallery_clip}
\end{subfigure}

\vskip 0.5\baselineskip

\begin{subfigure}[t]{\linewidth}
\centering
\includegraphics[width=\linewidth]{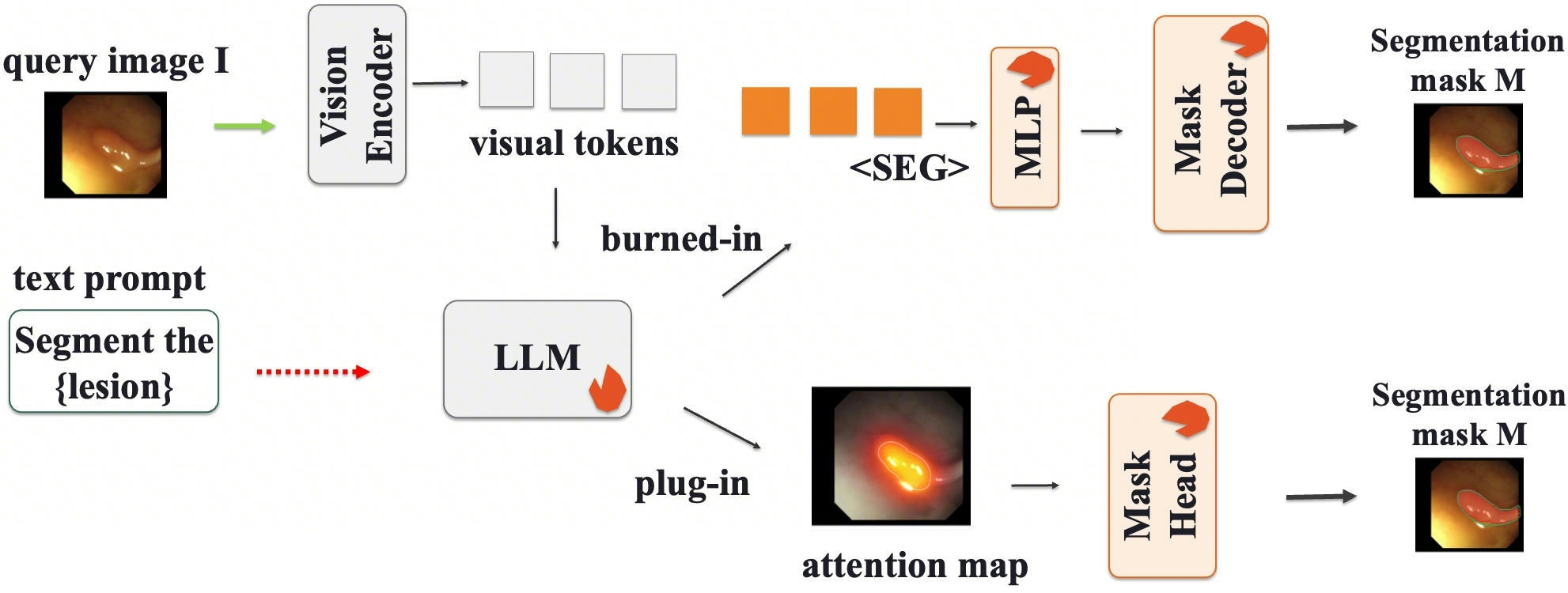}
\caption{MLLM-based.}
\label{fig:gallery_mllm}
\end{subfigure}

\vskip 0.5\baselineskip

\begin{subfigure}[t]{\linewidth}
\centering
\includegraphics[width=\linewidth]{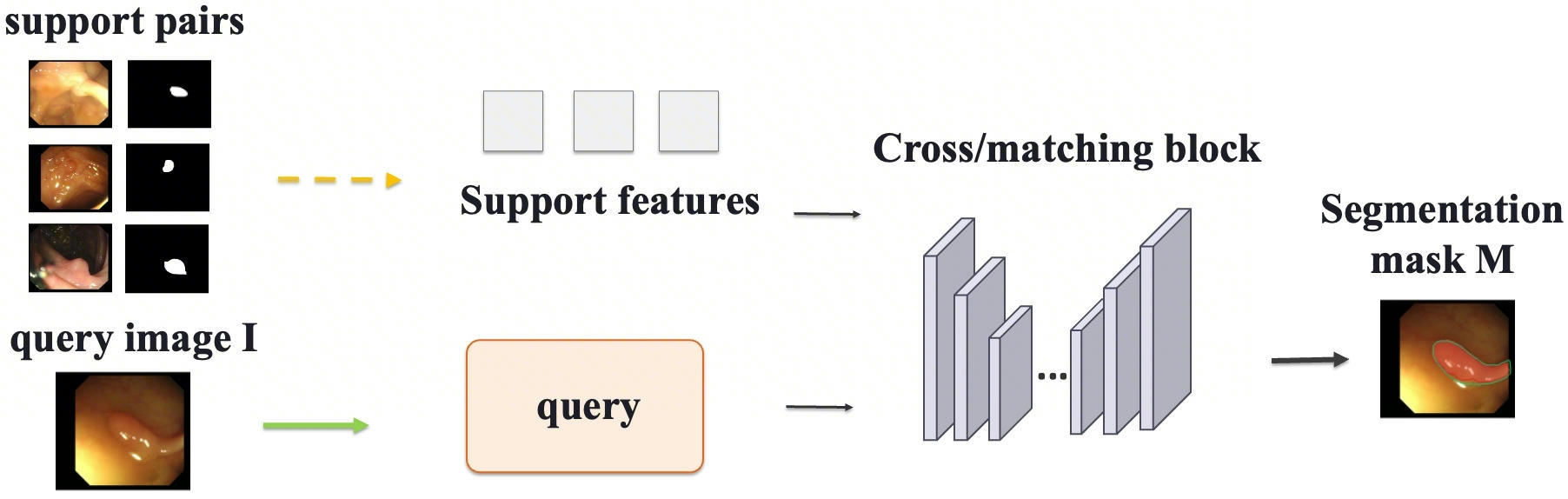}
\caption{Specialists.}
\label{fig:gallery_meta}
\end{subfigure}

\caption{Adaptation of each method family under \benchmark.}
\label{fig:gallery_methods}
\end{figure}

We group the evaluated methods into four solution categories: SAM-based methods, CLIP-based methods, MLLM-based methods, and specialists (see Tab.~\ref{tab:compared_methods}).Fig.~\ref{fig:gallery_methods} illustrates how each family adapts under \benchmark.


\subsection{SAM-based Methods}
We evaluate SAM~\cite{kirillov2023segment}, MedSAM~\cite{ma2024medsam}, MedSAM3~\cite{liu2025medsam3delvingsegmentmedical}, ProtoSAM~\cite{ayzenberg2025protosam}, and FoB-SAM~\cite{bo2026focus}, which differ in how support examples are ultilized.
SAM and MedSAM require spatial prompts at inference time but lack dedicated strategies for deriving them from support examples in FS-MIS. 
We therefore follow SAM-MPA~\cite{xu2024sammpa}, which avoids heuristic prompt selection by propagating support masks to query images and generating automatic spatial prompts from coarse predictions, obtaining SAM-MPA and MedSAM-MPA, respectively.
ProtoSAM and FoB-SAM are originally designed for one-shot inference. 
To evaluate them under 10-shot setting, we apply their official one-shot procedures independently to each support pair and aggregate predictions through post-processing, using probability averaging for ProtoSAM (denoted as ProtoSAM-post) and majority voting for FoB-SAM (denoted as FoB-SAM-post).


\subsection{CLIP-based Methods}
We evaluate SCLIP~\cite{wang2024scliprethinkingselfattentiondense}, MaskCLIP~\cite{zhou2022maskclip}, and ProxyCLIP~\cite{lan2024proxyclipproxyattentionimproves}. These methods derive dense predictions from the correspondence between visual features and a textual ROI description, without a task-specific mask decoder. In the zero-shot mode, each localizes the ROI using its original text-conditioned mechanism with a hand-crafted class name as the text input.
Since these methods are driven entirely by the text input, their segmentation quality depends heavily on how well that text describes the target, yet a fixed class name is often a poor description for medical ROIs. We therefore adapt them with CoOp~\cite{zhou2022coop}, which replaces the hand-crafted text with prompt tokens learned from the support examples. CoOp tunes exactly the text input these methods rely on while keeping both encoders frozen, giving a natural and parameter-efficient way to pass support information to this family. Our first configuration, CoOp, freezes both encoders and optimizes only the learnable text prompt tokens together with a learnable similarity-scaling parameter. The second, CoOp + LoRA, additionally applies LoRA to the visual encoder on top of the first configuration, so the dense representation can further specialize to the task.

\subsection{MLLM-based Methods}

We divide MLLM-based methods into \emph{burned-in MLLMs} and \emph{plug-in MLLMs}. Burned-in MLLMs already contain a pretrained text-to-mask mechanism in their released checkpoints, whereas plug-in MLLMs acquire segmentation capability by attaching and training an external segmentation interface while keeping the underlying MLLM frozen.
For reproducibility, we focus on open-source MLLMs with up to 8B.

\paragraph{Burned-in MLLMs.}
We carefully select representative burned-in MLLMs covering three mask-generation paradigms.
Embedding-based methods, including LISA~\cite{lai2024lisa}, READ~\cite{Qian_2025_CVPR}, GLaMM~\cite{rasheed2024glamm}, and MedPLIB~\cite{medplib}, represent masks through dedicated segmentation tokens, whose hidden states are mapped to external mask decoders. 
Autoregressive methods, represented by Text4Seg~\cite{lan2025text4seg}, represent masks as discrete token sequences generated autoregressively. 
All-mask prediction methods, represented by STAMP~\cite{liu2026stamp}, generate the complete mask-token sequence in parallel through a fill-in-the-blank formulation. 
For all burned-in MLLMs, we preserve their original mask representations and follow their official fine-tuning protocols when available.
MedPLIB~\cite{medplib} supports both LoRA-based fine-tuning and in-context inference. 
We denote the LoRA-adapted model as MedPLIB-LoRA. 
Since the released MedPLIB supports up to three in-context examples, we report its native 3-shot performance (MedPLIB-ICL) and a ten-shot variant, MedPLIB-post, which aggregates ten independent one-shot predictions.

\begin{table*}[!t]
\centering
\small
\renewcommand{\arraystretch}{0.6}
\resizebox{\textwidth}{!}{%
\begin{tabular}{l *{13}{c}}
\toprule
\multirow{2}{*}{\textbf{Method}}& \multicolumn{2}{c}{\textbf{Brain}} & \multicolumn{2}{c}{\textbf{Breast}} & \multicolumn{2}{c}{\textbf{Colon}} & \multicolumn{2}{c}{\textbf{Retina}} & \textbf{Lung} & \textbf{Skin} & \textbf{Thyroid} & \multicolumn{2}{c}{\textbf{Avg}} \\
\cmidrule(lr){2-3}\cmidrule(lr){4-5}\cmidrule(lr){6-7}\cmidrule(lr){8-9}\cmidrule(lr){13-14}
 & Dice & Spe. & Dice & Spe. & Dice & Spe. & Dice & Spe. & Dice & Dice & Dice & Dice & Spe. \\
\midrule
\multicolumn{14}{l}{\emph{Zero-shot mode}} \\
\midrule
MaskCLIP\fclip & 0.040 & 0.062 & 0.140 & 0.000 & 0.158 & 0.000 & 0.047 & 0.001 & 0.025 & 0.379 & 0.043 & 0.119 & 0.016 \\
LISA-7B\fmllm & 0.048 & \underline{0.302} & 0.188 & 0.000 & 0.133 & 0.000 & 0.113 & 0.001 & 0.022 & 0.193 & 0.243 & 0.134 & 0.076 \\
ProxyCLIP\fclip & 0.058 & 0.103 & 0.180 & 0.000 & 0.212 & 0.000 & 0.076 & 0.003 & 0.030 & 0.417 & 0.112 & 0.155 & 0.027 \\
SCLIP\fclip & 0.041 & 0.120 & 0.181 & 0.000 & 0.212 & 0.000 & 0.090 & 0.002 & 0.028 & 0.426 & 0.163 & 0.163 & 0.031 \\
Text4Seg\fmllm & 0.053 & 0.200 & 0.170 & \textbf{0.850} & 0.280 & \underline{0.872} & 0.082 & \underline{0.308} & 0.041 & 0.444 & 0.100 & 0.167 & \textbf{0.557} \\
STAMP-2B\fmllm & 0.063 & 0.000 & 0.321 & 0.000 & 0.294 & 0.000 & 0.122 & 0.000 & 0.070 & 0.582 & 0.293 & 0.249 & 0.000 \\
READ-7B\fmllm & 0.061 & \textbf{0.597} & 0.339 & \underline{0.023} & 0.343 & \textbf{0.992} & 0.066 & \textbf{0.498} & 0.056 & 0.681 & 0.278 & 0.261 & \underline{0.528} \\
STAMP-7B\fmllm & 0.069 & 0.100 & 0.516 & 0.000 & 0.357 & 0.000 & 0.114 & 0.000 & 0.062 & 0.673 & 0.308 & 0.300 & 0.025 \\
GLaMM-7B\fmllm & 0.065 & 0.000 & 0.432 & 0.000 & 0.382 & 0.000 & \textbf{0.130} & 0.000 & 0.052 & 0.695 & 0.343 & 0.300 & 0.000 \\
MedPLIB-7B\fmllm & \underline{0.187} & 0.000 & 0.509 & 0.000 & 0.254 & 0.000 & 0.100 & 0.000 & 0.096 & \underline{0.825} & 0.277 & 0.321 & 0.000 \\
MedSAM3\fsam & \textbf{0.385} & 0.091 & \textbf{0.768} & 0.000 & \textbf{0.798} & 0.000 & 0.121 & 0.011 & \textbf{0.680} & \textbf{0.894} & \textbf{0.645} & \textbf{0.613} & 0.026 \\
\midrule
\multicolumn{14}{l}{\emph{Few-shot mode}} \\
\midrule
FoB-SAM-post\fsam & 0.044 & 0.121 & 0.020 & 0.000 & 0.024 & 0.000 & 0.001 & 0.302 & 0.000 & 0.107 & 0.015 & 0.030 & 0.106 \\
FoB-SAM\fsam & 0.039 & 0.033 & 0.023 & 0.000 & 0.020 & 0.000 & 0.001 & 0.301 & 0.006 & 0.112 & 0.026 & 0.032 & 0.084 \\
Text4Seg\fmllm & 0.033 & 0.203 & 0.098 & \underline{0.880} & 0.138 & 0.857 & 0.047 & 0.393 & 0.021 & 0.235 & 0.071 & 0.092 & 0.583 \\
MedSAM-MPA\fsam & 0.009 & 0.096 & 0.268 & 0.000 & 0.132 & 0.023 & 0.048 & 0.109 & 0.019 & 0.349 & 0.177 & 0.143 & 0.057 \\
MedSAM\fsam & 0.060 & 0.000 & 0.227 & 0.000 & 0.213 & 0.000 & 0.110 & 0.000 & 0.025 & 0.413 & 0.229 & 0.182 & 0.000 \\
FAMNet\fs& 0.021 & 0.394 & 0.312 & 0.000 & 0.262 & 0.000 & 0.215 & 0.327 & 0.082 & 0.211 & 0.194 & 0.185 & 0.180 \\
LENS-Qwen2VL-2B\fmllm & 0.072 & 0.120 & 0.549 & 0.000 & 0.179 & 0.008 & 0.040 & 0.162 & 0.049 & 0.523 & 0.145 & 0.222 & 0.072 \\
LENS-LLaVA-1.5-7B\fmllm & 0.154 & 0.191 & 0.452 & 0.000 & 0.228 & 0.000 & 0.097 & 0.127 & 0.027 & 0.462 & 0.197 & 0.231 & 0.079 \\
F-LMM-MiniGemini-7B\fmllm & 0.000 & 0.000 & 0.481 & 0.000 & 0.159 & 0.000 & 0.028 & 0.000 & 0.002 & 0.659 & 0.312 & 0.234 & 0.000 \\
PANet\fs& 0.093 & 0.243 & 0.365 & 0.111 & 0.250 & 0.259 & 0.136 & \textbf{0.957} & 0.091 & 0.549 & 0.209 & 0.242 & 0.393 \\
LENS-LLaVA-OV\fmllm & 0.189 & 0.059 & 0.485 & 0.000 & 0.325 & 0.000 & 0.075 & 0.006 & 0.035 & 0.588 & 0.091 & 0.255 & 0.016 \\
F-LMM-LLaVA-1.5-7B\fmllm & 0.046 & 0.000 & 0.514 & 0.000 & 0.135 & 0.000 & 0.085 & 0.000 & 0.068 & 0.641 & 0.313 & 0.258 & 0.000 \\
SAM-MPA\fsam & 0.115 & 0.096 & 0.375 & 0.000 & 0.341 & 0.023 & 0.138 & 0.109 & 0.080 & 0.548 & 0.321 & 0.274 & 0.057 \\
F-LMM-LLaVA-Next-Vicuna-7B\fmllm & 0.155 & 0.000 & 0.529 & 0.000 & 0.424 & 0.000 & 0.047 & 0.000 & 0.009 & 0.662 & 0.261 & 0.298 & 0.000 \\
MedPLIB-7B-ICL\fmllm & 0.225 & 0.000 & 0.493 & 0.000 & 0.221 & 0.000 & 0.103 & 0.000 & 0.107 & 0.832 & 0.324 & 0.329 & 0.000 \\
MedPLIB-7B-post\fmllm & 0.148 & 0.000 & 0.473 & 0.203 & 0.300 & 0.000 & 0.107 & 0.000 & 0.097 & 0.851 & 0.332 & 0.330 & 0.051 \\
F-LMM-MiniGemini-2B\fmllm & 0.129 & 0.000 & 0.567 & 0.000 & 0.372 & 0.000 & 0.155 & 0.000 & 0.048 & 0.661 & 0.399 & 0.333 & 0.000 \\
STAMP-7B\fmllm & 0.004 & \underline{0.991} & 0.663 & 0.361 & 0.646 & 0.045 & 0.128 & 0.474 & 0.000 & 0.639 & 0.265 & 0.335 & 0.468 \\
F-LMM-DeepSeekVL-1.3B\fmllm & 0.103 & 0.000 & 0.452 & 0.000 & 0.551 & 0.000 & 0.145 & 0.000 & 0.137 & 0.653 & 0.350 & 0.341 & 0.000 \\
SCLIP-CoOp\fclip & 0.138 & 0.734 & 0.490 & 0.444 & 0.466 & 0.741 & 0.180 & 0.346 & 0.130 & 0.689 & 0.384 & 0.354 & 0.566 \\
LISA-7B\fmllm & 0.250 & 0.347 & 0.691 & 0.000 & 0.465 & 0.000 & 0.146 & 0.531 & 0.063 & 0.717 & 0.272 & 0.372 & 0.219 \\
\rowcolor{green!18}U-Net\fs & 0.405 & 0.426 & 0.306 & 0.000 & 0.330 & 0.000 & 0.234 & 0.359 & 0.207 & 0.703 & 0.431 & 0.374 & 0.196 \\
READ-7B\fmllm & 0.182 & 0.118 & 0.670 & 0.000 & 0.572 & 0.000 & 0.090 & 0.009 & 0.028 & 0.801 & 0.444 & 0.398 & 0.032 \\
ProtoSAM\fsam & 0.363 & 0.113 & 0.547 & 0.000 & 0.605 & 0.000 & 0.131 & 0.101 & 0.115 & 0.606 & 0.456 & 0.404 & 0.054 \\
RSRM\fs& 0.019 & \textbf{1.000} & 0.672 & 0.015 & \underline{0.670} & 0.000 & 0.099 & 0.333 & 0.000 & 0.815 & 0.601 & 0.411 & 0.337 \\
STAMP-2B\fmllm & 0.000 & \textbf{1.000} & 0.722 & 0.053 & 0.635 & 0.496 & 0.125 & 0.650 & 0.002 & \underline{0.853} & 0.615 & 0.422 & 0.550 \\
ProtoSAM-post\fsam & \underline{0.442} & 0.103 & 0.197 & \textbf{1.000} & 0.630 & 0.000 & 0.207 & 0.000 & \underline{0.367} & 0.728 & 0.383 & 0.422 & 0.276 \\
UniverSeg\fs& 0.351 & 0.361 & 0.604 & 0.000 & 0.398 & 0.000 & 0.166 & 0.001 & 0.212 & 0.739 & 0.547 & 0.431 & 0.091 \\
Tyche\fs& 0.287 & 0.407 & 0.674 & 0.083 & 0.399 & 0.150 & 0.189 & 0.123 & 0.107 & 0.798 & 0.572 & 0.432 & 0.191 \\
ProxyCLIP-CoOp\fclip & 0.282 & 0.755 & 0.610 & 0.444 & 0.552 & 0.481 & 0.231 & 0.391 & 0.116 & 0.828 & 0.450 & 0.438 & 0.518 \\
MaskCLIP-CoOp\fclip & 0.268 & 0.581 & 0.576 & 0.000 & 0.527 & 0.259 & 0.224 & 0.063 & 0.230 & 0.771 & 0.496 & 0.442 & 0.226 \\
MedPLIB-7B-LoRA\fmllm & 0.399 & 0.036 & 0.605 & 0.000 & 0.474 & 0.000 & 0.160 & 0.003 & 0.248 & 0.813 & 0.417 & 0.445 & 0.010 \\
SCLIP-LoRA\fclip & 0.165 & 0.933 & 0.598 & 0.778 & 0.583 & 0.852 & 0.230 & 0.385 & 0.256 & 0.829 & 0.560 & 0.460 & \underline{0.737} \\
ProxyCLIP-LoRA\fclip & 0.272 & 0.914 & 0.662 & 0.519 & 0.577 & \textbf{0.926} & 0.223 & \underline{0.662} & 0.115 & 0.835 & 0.571 & 0.465 & \textbf{0.755} \\
GLaMM-7B\fmllm & 0.269 & 0.375 & \underline{0.751} & 0.000 & 0.661 & 0.015 & 0.179 & 0.177 & 0.160 & 0.778 & 0.565 & 0.480 & 0.142 \\
MaskCLIP-LoRA\fclip & 0.263 & 0.905 & 0.651 & 0.481 & 0.595 & \underline{0.889} & \underline{0.238} & 0.435 & 0.217 & 0.838 & \underline{0.617} & \underline{0.488} & 0.677 \\
MedSAM3\fsam & \textbf{0.554} & 0.100 & \textbf{0.780} & 0.000 & \textbf{0.828} & 0.000 & \textbf{0.308} & 0.084 & \textbf{0.698} & \textbf{0.903} & \textbf{0.705} & \textbf{0.683} & 0.046 \\
\bottomrule
\end{tabular}%
}
\caption{Results on \benchmark per anatomy. The best value is \textbf{bold} and the second best is \underline{underlined}. Solution categories: \fs\,specialists, \fsam\,SAM-based, \fclip\,CLIP-based, \fmllm\,MLLM-based.}
\label{tab:fame_master}
\end{table*}

\paragraph{Plug-in MLLMs.}
We evaluate F-LMM~\cite{wu2025flmmgroundingfrozenlarge} and LENS~\cite{liu2025segmentationplugandplaycapabilityfrozen} as representative plug-in MLLMs. Both keep the pretrained MLLM frozen and add a trainable segmentation interface that converts its visual-language representations into dense masks. We follow their official implementations and adaptation protocols, training only the designated segmentation components.

\subsection{Specialists}

Specialists are segmentation models purpose-built for medical image segmentation. U-Net~\cite{ronneberger2015unet} is trained from scratch on target-task annotations, updating all its parameters. The remaining specialists---PANet~\cite{wang2019panet}, UniverSeg~\cite{butoi2023universeg}, Tyche~\cite{rakic2024tyche}, FAMNet~\cite{bo2025famnet}, and RSRM~\cite{sun2026training}---segment query images by conditioning on annotated support image--mask pairs without task-specific parameter updates, transferring task information through prototype matching, support--query conditioning, frequency-domain matching, or feature correspondence.

\section{Experiments}

\subsection{Set up}
All training and inference are conducted on 8 RTX 3090 GPUs. For each method, we follow its native input format and adaptation protocol whenever available; otherwise, we apply the adaptation strategies specified in Tab.~\ref{tab:compared_methods}. 
Implementation details are provided in the supplementary material.




\begin{figure}[t]
\centering
\includegraphics[width=\linewidth]{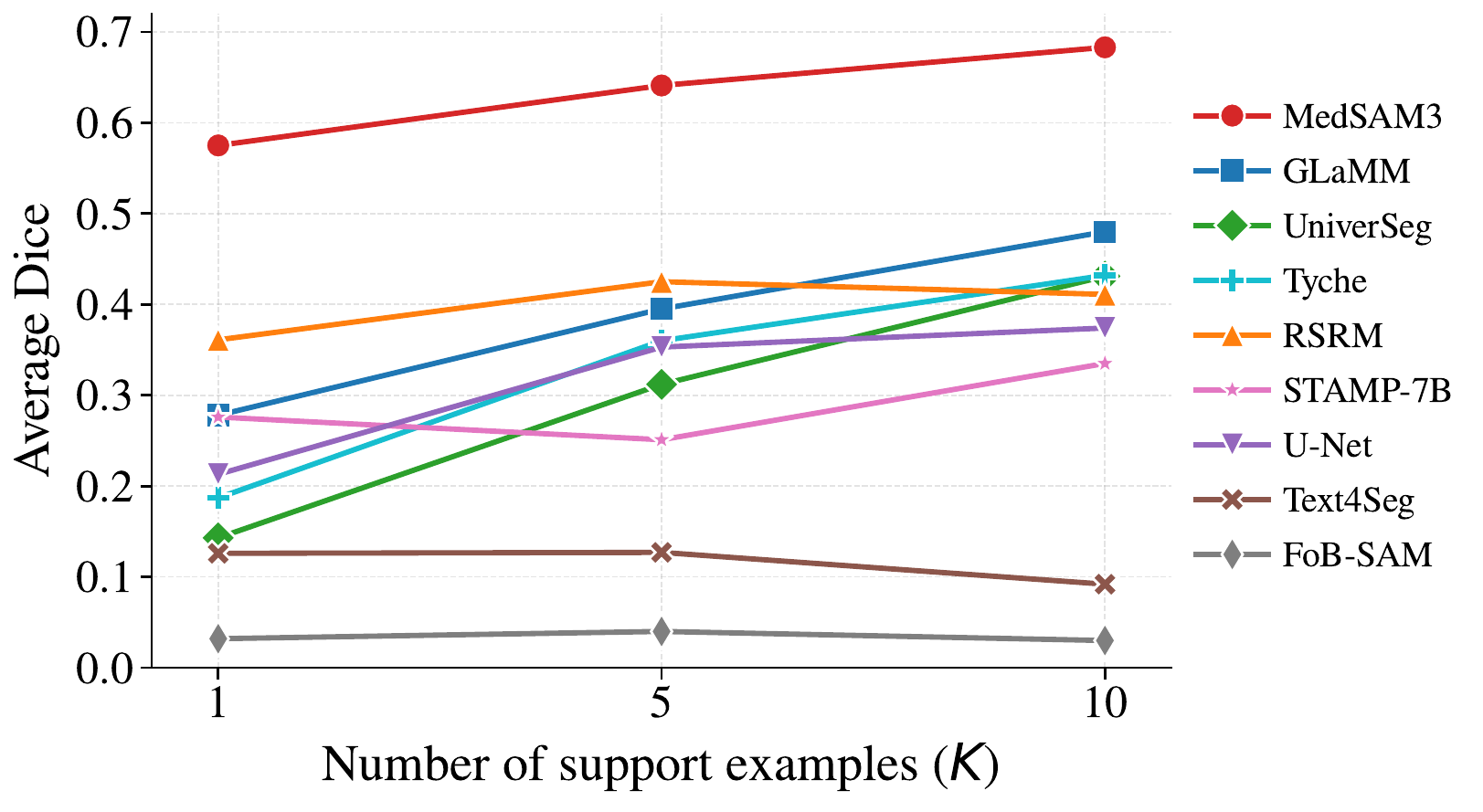}
\caption{Average Dice under varying support set size $K$.}
\label{fig:kshot_trend}
\end{figure}

\begin{table}[t]
\centering
\setlength{\tabcolsep}{3pt}
\renewcommand{\arraystretch}{1}
\resizebox{\columnwidth}{!}{%
\begin{tabular}{l ccccc}
\toprule
& \textbf{Covariate} & \multicolumn{2}{c}{\textbf{Semantic}} & \multicolumn{2}{c}{\textbf{Average}} \\
\cmidrule(lr){2-2}\cmidrule(lr){3-4}\cmidrule(lr){5-6}
\textbf{Method} & Dice & Dice & Spe. & Dice & Spe. \\
\midrule
Text4Seg\fmllm & 0.148 & 0.051 & 0.202 & 0.100 & 0.202 \\
MedSAM-MPA\fsam & 0.152 & 0.075 & 0.013 & 0.114 & 0.013 \\
LENS-LLaVA-1.5-7B\fmllm & 0.196 & 0.063 & 0.092 & 0.129 & 0.092 \\
LENS-Qwen2VL-2B\fmllm & 0.208 & 0.062 & 0.067 & 0.135 & 0.067 \\
LENS-LLaVA-OV\fmllm & 0.236 & 0.049 & 0.067 & 0.143 & 0.067 \\
MedSAM\fsam & 0.199 & 0.101 & 0.000 & 0.150 & 0.000 \\
\rowcolor{green!18}U-Net\fs & 0.263 & 0.101 & 0.467 & 0.182 & 0.467 \\
F-LMM-LLaVA-Next-Vicuna-7B\fmllm & 0.323 & 0.078 & 0.000 & 0.200 & 0.000 \\
F-LMM-MiniGemini-7B\fmllm & 0.307 & 0.100 & 0.000 & 0.203 & 0.000 \\
F-LMM-LLaVA-1.5-7B\fmllm & 0.312 & 0.114 & 0.000 & 0.213 & 0.000 \\
SAM-MPA\fsam & 0.309 & 0.129 & 0.013 & 0.219 & 0.013 \\
F-LMM-DeepSeekVL-1.3B\fmllm & 0.356 & 0.088 & 0.000 & 0.222 & 0.000 \\
MedPLIB-7B-ICL\fmllm & 0.325 & 0.124 & 0.000 & 0.224 & 0.000 \\
LISA-7B\fmllm & 0.391 & 0.087 & 0.372 & 0.239 & 0.372 \\
GLaMM\fmllm & 0.341 & 0.150 & 0.000 & 0.245 & 0.000 \\
F-LMM-MiniGemini-2B\fmllm & 0.380 & 0.111 & 0.002 & 0.246 & 0.002 \\
SCLIP-CoOp\fclip & 0.417 & 0.092 & 0.493 & 0.255 & 0.493 \\
STAMP-7B\fmllm & 0.435 & 0.077 & \textbf{0.837} & 0.256 & \textbf{0.837} \\
STAMP-2B\fmllm & 0.431 & 0.082 & \underline{0.807} & 0.257 & \underline{0.807} \\
MedPLIB-7B-post\fmllm & 0.373 & 0.149 & 0.000 & 0.261 & 0.000 \\
READ-7B\fmllm & 0.436 & 0.140 & 0.000 & 0.288 & 0.000 \\
MedPLIB-7B-LoRA\fmllm & 0.441 & \underline{0.172} & 0.005 & 0.306 & 0.005 \\
SCLIP-LoRA\fclip & 0.546 & 0.065 & 0.363 & 0.306 & 0.363 \\
MaskCLIP-LoRA\fclip & 0.543 & 0.077 & 0.431 & 0.310 & 0.431 \\
ProxyCLIP-LoRA\fclip & \underline{0.583} & 0.056 & 0.600 & 0.320 & 0.600 \\
ProxyCLIP-CoOp\fclip & 0.542 & 0.098 & 0.385 & 0.320 & 0.385 \\
MaskCLIP-CoOp\fclip & 0.532 & 0.155 & 0.269 & \underline{0.343} & 0.269 \\
MedSAM3\fsam & \textbf{0.677} & \textbf{0.362} & 0.067 & \textbf{0.519} & 0.067 \\
\bottomrule
\end{tabular}%
}
\caption{\textbf{Out-of-distribution results}. Rows are ordered by Average Dice from low to high.}
\label{tab:ood_results}
\end{table}

\subsection{Architecture and Adaptation Strategies}
Tab.~\ref{tab:fame_master} shows that effective few-shot segmentation depends more on how models exploit annotated support examples than on model scale alone. 
Across architectures, methods that directly adapt visual representations or mask generation generally outperform those relying on indirect semantic guidance.
Among SAM-based methods, MedSAM3 achieves the best performance (0.683 Dice), substantially outperforming SAM-MPA and MedSAM-MPA (0.274 and 0.143). 
This gap suggests that converting support examples into spatial prompts may lose fine-grained localization cues, whereas dense representation adaptation enables support masks to directly guide pixel-level prediction.
A similar trend is observed for CLIP-based methods. 
LoRA consistently improves over CoOp for SCLIP (0.354$\rightarrow$0.460) and ProxyCLIP (0.438$\rightarrow$0.465), indicating that adapting visual features is more effective than refining only textual semantics for medical segmentation.
MLLM-based methods further show that mask representation is more important than language model scale. 
Methods with direct visual-to-mask generation, such as GLaMM and MedPLIB, generally outperform lightweight plug-in approaches. 
In contrast, Text4Seg underperforms because its intermediate mask representation relies on discrete language tokens, introducing a bottleneck for pixel-level prediction. 
Moreover, larger language models do not consistently improve segmentation, as STAMP-2B outperforms STAMP-7B and F-LMM-MiniGemini-2B outperforms its 7B counterpart.

\subsection{Scaling with Support Examples}

Fig.~\ref{fig:kshot_trend} shows that increasing the number of support examples improves performance only when models can effectively utilize them. 
Models with stronger support utilization consistently benefit from larger support sets, including MedSAM3 (0.574$\rightarrow$0.683) and GLaMM (0.278$\rightarrow$0.480) from 1-shot to 10-shot.
In contrast, increasing support examples provides limited or even negative gains for models with weaker support utilization, such as Text4Seg (0.127$\rightarrow$0.092). 
These results indicate that the key bottleneck of few-shot segmentation is not the number of support examples, but whether models can effectively leverage support masks for pixel-level prediction.

\subsection{Generalization under Distribution Shifts}

Tab.~\ref{tab:ood_results} tests generalization under covariate and semantic shifts, and shows that moving to a new ROI is far harder than adapting across imaging domains. Semantic shift hurts much more than covariate shift: MedSAM3 drops from 0.677 to 0.362 Dice and the CLIP variants fall from above 0.5 to below 0.16. The best in-distribution adaptation can even backfire once the target changes: LoRA beats CoOp on covariate transfer for every CLIP backbone (e.g., ProxyCLIP: 0.583 vs. 0.542 Dice), but this order flips under semantic shift, where CoOp matches or beats LoRA (e.g., MaskCLIP: 0.155 vs. 0.077 Dice), showing that tuning the visual encoder to the source helps only while the target stays the same.

Segmenting a target and rejecting an absent one remain different skills: MedSAM3 stays the strongest segmenter (0.519 average Dice) but reaches only 0.067 specificity, while STAMP-7B and STAMP-2B reach 0.837 and 0.807 specificity with much lower Dice. 

Overall, \benchmark reveals that few-shot medical segmentation is constrained by two challenges: effectively transferring target-specific visual concepts and reliably recognizing target absence under distribution shifts.

\section{Conclusions}

We present \benchmark, a solution-level benchmark for few-shot medical image segmentation, covering specialists, SAM-based methods, CLIP-based methods, and MLLM-based methods under a unified evaluation protocol. 
Our evaluation reveals three key findings: (1) effective few-shot segmentation depends on how models exploit annotated support examples, with direct support-driven visual adaptation generally outperforming prompt or semantic adaptation; (2) increasing support examples improves performance only when models can effectively utilize them; and (3) semantic transfer remains a major challenge under distribution shifts, while segmentation accuracy does not necessarily imply reliable target-absence recognition. 
These findings motivate future FS-MIS methods that better leverage support examples for accurate and robust segmentation.

\textbf{Limitations.}
We evaluate open-source MLLMs up to 8B parameters and leave larger-scale models and improved adaptation mechanisms for future study.

\bibliography{aaai2027}
\end{document}